\documentclass[10pt,twocolumn,letterpaper]{article}

\usepackage{iccv}
\usepackage{times}
\usepackage{epsfig}
\usepackage{graphicx}
\usepackage{amsmath}
\usepackage{amssymb}
\usepackage[table,xcdraw]{xcolor}
\usepackage{multirow}
\usepackage{caption}
\usepackage{adjustbox}
\usepackage{tabularx}
\usepackage{booktabs}
\usepackage{array}
\usepackage{cuted}
\usepackage{subfig}
\def\code#1{\texttt{#1}}
\usepackage{amsfonts}
\let\oldsmallfrown\smallfrown
\renewcommand{\smallfrown}[1][0pt]{%
  \mathrel{\raisebox{#1}{$\oldsmallfrown$}}%
}

\definecolor{citecolor}{RGB}{9,113,188}
\newcolumntype{C}{>{\centering\arraybackslash}p{1.29cm}}

\usepackage[citecolor=citecolor,breaklinks=true,bookmarks=false,colorlinks]{hyperref}

\newcommand\rurl[1]{%
  \href{http://#1}{\nolinkurl{#1}}%
}

\iccvfinalcopy 

\def\iccvPaperID{****} 

\ificcvfinal\fi

\begin{document}

\title{High-Fidelity Pluralistic Image Completion with Transformers}

\author{Ziyu Wan$^{1}$ \quad \quad Jingbo Zhang$^{1}$ \quad \quad Dongdong Chen$^2$ \quad \quad Jing Liao$^{1}$\thanks{Corresponding author.} \\
    \\
	$^1$City University of Hong Kong \quad  
	$^2$Microsoft Cloud + AI
	\\
	\rurl{raywzy.com/ICT}
	}

\maketitle
\pagestyle{empty}  
\thispagestyle{empty} 
\begin{strip}
\vspace{-5.5em}
\setlength\tabcolsep{0.5pt}
\centering
\begin{tabular}{c}
    \vspace{-3.0mm}
    \includegraphics[width=1.0\textwidth]{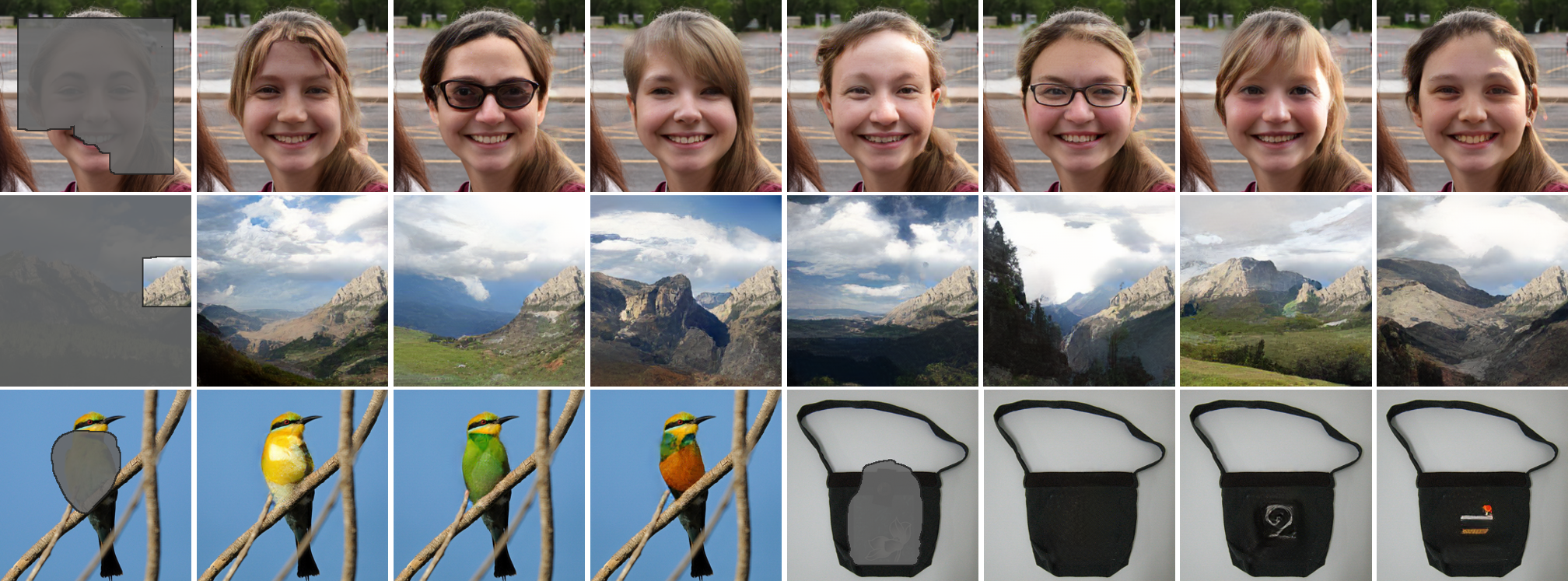}
     \end{tabular}
\captionof{figure}{Pluralistic free-form image completion results produced by our method.
}
\label{fig:teaser}
\end{strip}

\begin{abstract}
\vspace{-1em}
Image completion has made tremendous progress with convolutional neural networks (CNNs), because of their powerful texture modeling capacity. However, due to some inherent properties (\eg, local inductive prior, spatial-invariant kernels), CNNs do not perform well in understanding global structures or naturally support pluralistic completion. Recently, transformers demonstrate their power in modeling the long-term relationship and generating diverse results, but their computation complexity is quadratic to input length, thus hampering the application in processing high-resolution images. This paper brings the best of both worlds to pluralistic image completion: appearance prior reconstruction with transformer and texture replenishment with CNN. The former transformer recovers pluralistic coherent structures together with some coarse textures, while the latter CNN enhances the local texture details of coarse priors guided by the high-resolution masked images. The proposed method vastly outperforms state-of-the-art methods in terms of three aspects: 1) large performance boost on image fidelity even compared to deterministic completion methods; 2) better diversity and higher fidelity for pluralistic completion; 3) exceptional generalization ability on large masks and generic dataset, like ImageNet.

\end{abstract}

\section{Introduction}

Image completion (a.k.a. image inpainting), which aims to fill the missing parts of images with visually realistic and semantically appropriate contents, has been a longstanding and critical problem in computer vision areas. It is widely used in a broad range of applications, such as object removal~\cite{barnes2009patchmatch}, photo restoration~\cite{wan2020bringing,wan2020old}, image manipulation~\cite{jo2019sc}, and image re-targeting~\cite{cho2017weakly}. To solve this challenging task, traditional methods like PatchMatch~\cite{barnes2009patchmatch} usually search for similar patches within the image and paste them into the missing regions, but they require appropriate information to be contained in the input image, \eg, similar structures or patches, which is often difficult to satisfy. 

In recent years, CNN-based solutions~\cite{Iizuka-sig17-lgmatch4,Li-cvpr17-facecomplete5,Pathak-cvpr16-featinpaint7,Yu-cvpr18-attentioninpainting8, liu2019coherent} started to dominate this field. By training on large-scale datasets in a self-supervised way, CNNs have shown their strength in learning rich texture patterns, and fills the missing regions with such learned patterns. Besides, CNN models are computationally efficient considering the sparse connectivity of convolutions. Nonetheless, they share some inherent limitations: 1) The local inductive priors of convolution operation make modeling the global structures of an image difficult; 2) CNN filters are spatial-invariant, \ie the same convolution kernel operates on the features across all positions, by that the duplicated patterns or blurry artifacts frequently appear in the masked regions. On the other hand, CNN models are inherently deterministic. To achieve diverse completion outputs, recent frameworks~\cite{zheng2019pluralistic,zhao2020uctgan} rely on optimizing the variational lower bound of instance likelihood. Nonetheless, extra distribution assumption would inevitably hurt the quality of generated contents~\cite{zhao2017towards}.

Transformer, as well-explored architectures in language tasks, is on-the-rise in many computer vision tasks. Compared to CNN models, it abandons the baked-in local inductive prior and is designed to support the long-term interaction via the dense attention module~\cite{vaswani2017attention}.  Some preliminary works \cite{chen2020generative} also demonstrate its capacity in modeling the structural relationships for natural image synthesis. Another advantage of using a transformer for synthesis is that it naturally supports pluralistic outputs by directly optimizing the underlying data distribution. However, the transformer also has its own deficiency. Due to quadratically increased computational complexity with input length, it struggles in high-resolution image synthesis or processing. Besides, most existing transformer-based generative models ~\cite{parmar2018image, chen2020generative} works in an auto-regressive manner, \ie, synthesize pixels in a fixed order, like the raster-scan order, thus hampering its application in the image completion task where the missing regions are often with arbitrary shapes and sizes.

In this paper, we propose a new high-fidelity pluralistic image completion method by bringing the best of both worlds: the global structural understanding ability and pluralism support of transformer, and the local texture refinement ability and efficiency of CNN models. To achieve this, we decouple image completion into two steps: pluralistic appearance priors reconstruction with a transformer to recover the coherent image structures, and low-resolution upsampling with CNN to replenish fine textures. Specifically, given an input image with missing regions, we first leverage the transformer to sample low-resolution completion results, \ie appearance priors. Then, guided by the appearance priors and the available pixels of the input image, another upsampling CNN model is utilized to render high-fidelity textures for missing regions meanwhile ensuring coherence with neighboring pixels. In particular, unlike previous auto-regressive methods ~\cite{chen2020generative,van2016conditional}, in order to make the transformer model capable of completing the missing regions by considering all the available contexts, we optimize the log-likelihood objective of missing pixels based on the bi-directional conditions, which is inspired by the masked language model like BERT~\cite{devlin2018bert}.

To demonstrate the superiority, we compare our method with state-of-the-art deterministic~\cite{nazeri2019edgeconnect,liu2020rethinking,yu2019free} and pluralistic~\cite{zheng2019pluralistic} image completion approaches on multiple datasets. Our method makes significant progress from three aspects: 1) Compared with previous deterministic completion methods, our method establishes a new state of the art and outperforms theirs on a variety of metrics by a large margin; 2) Compared with previous pluralistic completion methods, our method further enhances the results diversity, meanwhile achieving higher completion fidelity; 3) Thanks to the strong structure modeling capacity of transformers, our method generalizes much better in completing extremely large missing region and large-scale generic datasets (\eg, ImageNet) as shown in Figure.~\ref{fig:teaser}. Remarkably, the FID score on ImageNet is improved by 41.2 at most compared with the state-of-the-art method PIC~\cite{zheng2019pluralistic}.

\section{Related Works}

\noindent\textbf{Visual Transformers} ~~ Vaswani \emph{et al.}~\cite{vaswani2017attention} firstly propose transformers for machine translation, whose success subsequently has been proved in various down-stream natural language processing (NLP) tasks. The overall architecture of transformers is composed of stacked self-attention and point-wise feed-forward layers for encoder and decoder. Since the attention mechanism could model the dense relationship among elements of input sequence well, transformers are now gradually popular in computer vision areas. For example, DETR~\cite{carion2020end} employ transformers as the backbone to solve the object detection problem. Dosovitskiy \emph{et al.}~\cite{dosovitskiy2020image} propose ViT, which firstly utilize transformers in the image recognition area and achieved excellent results compared with CNN-based methods. Besides, Parmar \emph{et al.}~\cite{parmar2018image} and Chen \emph{et al.}~\cite{chen2020generative} leverage the transformer to model the image. Nonetheless, to generate an image, these methods rely on a fixed permutation order, which is not suitable to complete the missing areas with varying shapes.

\begin{figure*}[!t]
\begin{center}
\includegraphics[width=1.0\linewidth]{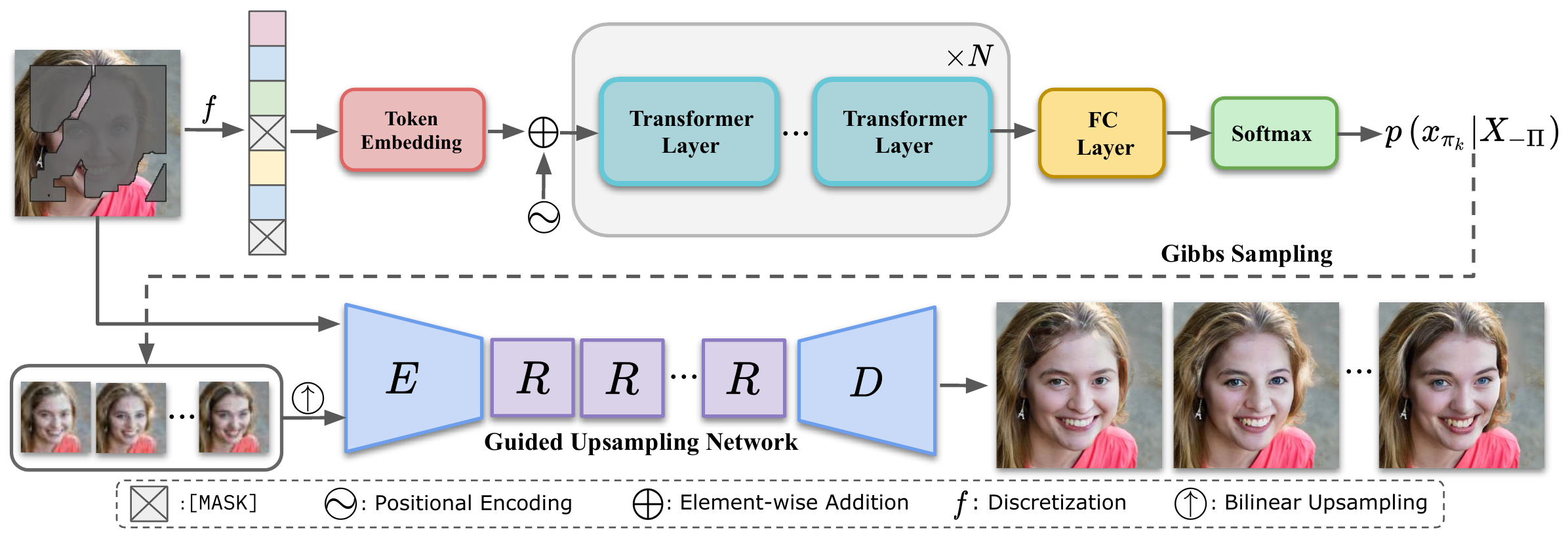}
\vspace{-1.4em}
\caption{\textbf{Pipeline Overview.} Our method consists of two networks. The above one is \textbf{bi-directional transformer}, which is responsible for  producing the probability distribution of missing regions, then the appearance priors could be reconstructed by sampling from this distribution with diversities. Subsequently, we employ another \textbf{CNN} to upsample the appearance prior to original resolution under the guidance of input masked images. Our method combines both advantages of transformer and CNN, leading to high-fidelity pluralistic image completion performance. \textbf{E}: Encoder, \textbf{D}: Decoder, \textbf{R}: Residual block.}
\label{fig:pipeline}
\end{center}
\vspace{-1.5em}
\end{figure*}

\noindent\textbf{Deterministic Image Completion} ~~ Traditional image completion methods, like diffusion-based~\cite{bertalmio2000image,efros2001image} and patch-based~\cite{barnes2009patchmatch, hays2007scene, darabi2012image}, rely on strong low-level assumptions,  which may be violated while facing large-area masks. To generate semantic-coherent contents, recently many CNN-based methods~\cite{Pathak-cvpr16-featinpaint7,liu2018image,liu2019coherent,Yu-cvpr18-attentioninpainting8,nazeri2019edgeconnect} have been proposed. Most of the methods share a similar encoder-decoder architecture. Specifically, Pathak \emph{et al.}~\cite{Pathak-cvpr16-featinpaint7} bring the adversarial training into inpainting and achieve semantic hole-filling. Iizuka \emph{et al.}~\cite{Iizuka-sig17-lgmatch4} improve the performance of CE~\cite{Pathak-cvpr16-featinpaint7} by involving a local-global discriminator. Yu \emph{et al.}~\cite{Yu-cvpr18-attentioninpainting8} propose a new contextual attention module to capture the long-range correlations. Liu \emph{et al.}~\cite{liu2018image} design a new operator named partial-conv to alleviate the negative influence of the masked regions produced by convolutions. These methods could generate reasonable contents for masked regions but lack the ability to generate diversified results.

\noindent\textbf{Pluralistic Image Completion} ~~ To obtain a diverse set of results for each masked input, Zheng \emph{et al.}~\cite{zheng2019pluralistic} propose a dual pipeline framework, which couples the estimated distribution from the reconstructive path and conditional prior of the generative path via jointly maximizing the lower bound. Similar with \cite{zheng2019pluralistic}, UCTGAN~\cite{zhao2020uctgan} project both the masked input and reference image into a common space via optimizing the KL-divergence between encoded features and $\mathcal{N}(\mathbf{0}, \mathbf{I})$ distribution to achieve diversified sampling. Although they have achieved some diversities to a certain extent, their completion qualities are limited due to variational training. Unlike these methods, we directly optimize the log-likelihood in the discrete space via transformers without auxiliary assumptions.

\section{Method}
Image completion aims to transform the input image $I_m\in \mathbb{R}^{H \times W \times 3}$ with missing pixels into a complete image $I\in \mathbb{R}^{H \times W \times 3}$. This task is inherently stochastic, which means given the masked image $I_m$, there exists a conditional distribution $p(I| I_m)$. We decompose the completion procedure into two steps, appearance priors $X$ reconstruction and texture details replenishment. Since obtaining coarse prior $X$ given $I$ and $I_m$ is deterministic, then $p(I| I_m)$ could be re-written as,
\begin{equation}
\begin{split}
    p(I| I_m)=&p(I| I_m)\cdot p(X| I,I_m) \\
    =&p(X,I | I_m ) \\
    =&p(X| I_m)\cdot p(I| X,I_m).
\end{split}
\end{equation}
Instead of directly sampling from $p(I| I_m)$, we first use a transformer to model the underlying distribution of appearance priors given $I_m$, denoted as $p(X| I_m)$ (described in Sec.~\ref{sec3.1}). These reconstructed appearance priors contain ample cues of global structure and coarse textures, thanks to the transformer's strong representation ability. Subsequently, we employ another CNN to replenish texture details under the guidance of appearance priors and unmasked pixels, denoted as $p(I| X,I_m)$ (described in Sec.~\ref{sec3.2}). The overall pipeline could be found in Figure.~\ref{fig:pipeline}.

\subsection{Appearance Priors Reconstruction}\label{sec3.1}

\noindent\textbf{Discretization} ~~ Considering the quadratically increasing computational cost of multi-head attention~\cite{vaswani2017attention} in the transformer architecture, we represent the appearance priors of a natural image with its corresponding low-resolution version ($32\times32$ or $48\times48$ in our implementation), which contains structural information and coarse textures only. Nonetheless, the dimensionality of RGB pixel representation ($256^3$) is still too large. To further reduce the dimension and faithfully re-represent the low-resolution image, an extra visual vocabulary with spatial size $512\times3$ is generated using K-Means cluster centers of the whole ImageNet~\cite{deng2009imagenet} RGB pixel spaces. Then for each pixel of appearance priors, we search the index of the nearest element from the visual vocabulary to obtain its discrete representation. In addition, the elements of the representation sequence corresponding to hole regions will be replaced with a special token \code{[MASK]}, which is also the learning target of the transformer. To this end, we convert the appearance prior into a discretized sequence.

\noindent\textbf{Transformer} ~~  For each token of the discretized sequence $X=\left\{x_1,x_2, \cdots, x_{\mathbb{L}}\right\}$, where $\mathbb{L}$ is the length of $X$, we project it into a $d-$dimensional feature vector through prepending a learnable embedding. To encode the spatial information, extra learnable position embeddings will be added into the token features for every location $1 \leq  i \leq  \mathbb{L}$ to form the final input $E\in \mathbb{R}^{\mathbb{L} \times d}$ of transformer model.

\begin{figure}[!t]
\begin{center}
\includegraphics[width=1.0\linewidth]{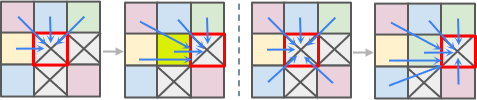}
\vspace{-1.4em}
\caption{Differences between single-directional (left) and bi-directional (right) attention.}
\label{fig:sampling}
\end{center}
\vspace{-2em}
\end{figure}

Following GPT-2~\cite{radford2019language}, we use decoder-only transformer as our network architecture, which is mainly composed with $N$ self-attention based transformer layers. At each transformer layer $l$, the calculation could be formulated as 
\vspace{-0.1cm}
\begin{equation}
\begin{split}
    F^{l-1}=&\text{LN}(\text{MSA}\left(E^{l-1}\right))+E^{l-1} \\
    E^{l}=&\text{LN}(\text{MLP}\left(F^{l-1}\right))+F^{l-1},
\end{split}
\end{equation}
where LN, MSA, MLP denote layer normalization~\cite{ba2016layer}, multi-head self-attention and FC layer respectively. More specifically, given input $E$, the MSA could be computed as:
\begin{equation}
\begin{split}
    head_{j}=&\operatorname{softmax}\left(\frac{E\mathbf{W}^{j}_{Q}(E\mathbf{W}_{K}^{j})^{T}}{\sqrt{d}}\right)(E\mathbf{W}_{V}^{j})\\
    \text{MSA}=&[head_1;...;head_h]\mathbf{W}_{O},
\end{split}
\end{equation}
where $h$ is the number of head, $\mathbf{W}_{Q}^{j}$, $\mathbf{W}_{K}^{j}$ and $\mathbf{W}_{V}^{j}$ are three learnable linear projection layers, $1 \leq  j \leq  h$. $\mathbf{W}_{O}$ is also a learnable FC layer, whose target is to fuse the concatenation of the outputs from different heads. By adjusting the parameters of transformer layer $N$, embedding dimension $d$ and head number $h$, we could easily scale the size of the transformer. It should also be noted that unlike auto-regressive transformers ~\cite{chen2020generative,parmar2018image}, which generate elements via single-directional attention, \ie only constrained by the context before scanning line,  we make each token attend to all positions to achieve bi-directional attention, as shown in Figure.~\ref{fig:sampling}. This ensures the generated distribution could capture all available contexts, either before and after a mask position in the raster-scan order, thus leading to the consistency between generated contents and unmasked regions. 

The output of the final transformer layer is further projected to a per-element distribution over 512 elements of visual vocabulary with the fully connected layers and softmax function. We adopt the masked language model (MLM) objective similar as the one used in BERT~\cite{devlin2018bert} to optimize the transformer model. Specifically, let $\Pi=\left\{\pi_{1}, \pi_{2}, \ldots, \pi_{K}\right\}$ denote the indexes of \code{[MASK]} tokens in the discretized input, where $K$ is the number of masked tokens. Let $X_{\Pi}$ denote the set of \code{[MASK]} tokens in $X$, and $X_{-\Pi}$ denote the set of unmasked tokens. The objective of MLM minimizes
the negative log-likelihood of $X_{\Pi}$ conditioned on all observed regions:
\begin{equation}
    L_{MLM}=\underset{X }{\mathbb{E}} [\frac{1}{K} \sum_{k=1}^{K}- \log p\left(x_{\pi_{k}}|X_{-\Pi},\theta \right)],
\end{equation}
where $\theta$ is the parameters of transformer. MLM objective incorporating with bi-directional attention ensures that the transformer model could capture the whole contextual information to predict the probability distribution of missing regions.

\begin{figure*}[t!]
    \begin{center}
    \includegraphics[width=1.0\linewidth]{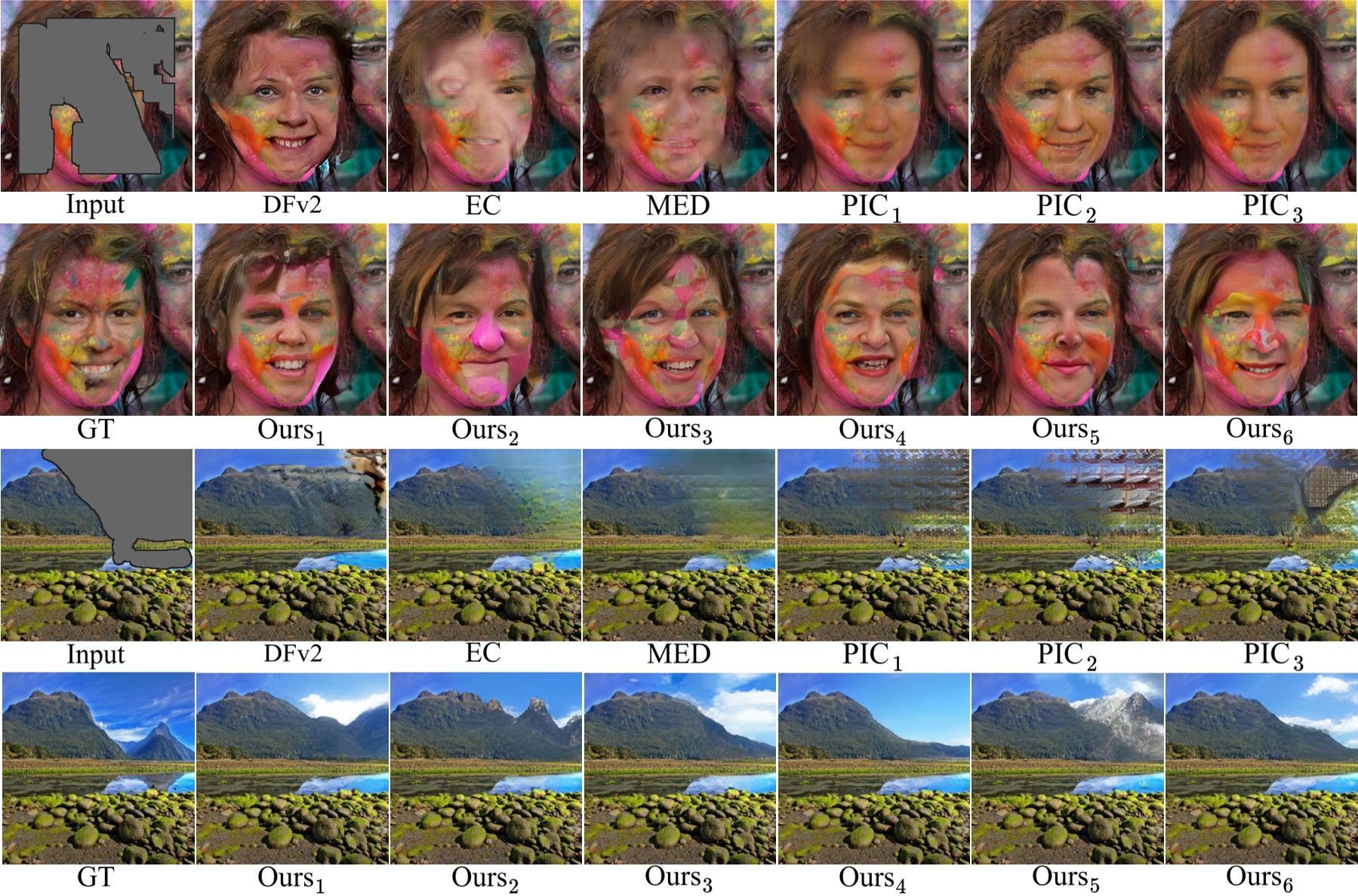}
    \end{center}
    \vspace{-1.4em}
    \caption{\textbf{Qualitative comparison with state-of-the-art methods on FFHQ, Places2 dataset.} The completion results of our method are with better quality and diversity.}
    \label{figure:qualititative_face_place}
    \vspace{-0.8em}
\end{figure*}

\noindent\textbf{Sampling Strategy} ~~   We introduce how to obtain reasonable and diversified appearance priors using the trained transformer in this section. Given the generated distribution of the transformer, directly sampling the entire set of masked positions does not produce good results due to the independence property. Instead, we employ Gibbs sampling to iteratively sample tokens at different locations. Specifically, in each iteration, a grid position is sampled from
$p\left(x_{\pi_{k}}|X_{-\Pi},X_{<\pi_{k}},\theta \right)$ with the top-$\mathcal{K}$ predicted elements, where $X_{<\pi_{k}}$ denotes the previous generated tokens. Then the corresponding \code{[MASK]} token is replaced with the
sampled one, and the process is repeated until all
positions are updated. Similar with PixelCNN~\cite{van2016conditional}, the positions are sequentially chosen in a raster-scan manner by default. After sampling, we could obtain a bunch of complete token sequences. For each complete discrete sequences sampled from transformer, we reconstruct its appearance priors $X \in \mathbb{R} ^{ \mathbb{L} \times 3} $ with querying the visual vocabulary.

\subsection{Guided Upsampling}~\label{sec3.2}
After reconstructing the low-dimensional appearance priors, we reshape $X$ into $I_t \in \mathbb{R}^{\sqrt{\mathbb{L}} \times \sqrt{\mathbb{L}} \times 3}$ for subsequent processing. Since $I_t$ has contained the diversity, now the problem is how to learn a deterministic mapping to re-scale the $I_t$ into original resolution $H\times W \times3$, meanwhile preserving the boundary consistency between hole regions and unmasked regions. To achieve this point, since CNNs have advantages of modeling texture patterns, here we introduce another guided upsampling network, which could render high-fidelity details of reconstructed appearance priors with the guidance of masked input $I_m$. The processing of guided upsampling could be written as 
\begin{equation}
    I_{pred}= \mathcal{F}_{\delta}(I_t^{\uparrow}\smallfrown[4pt] I_m)\in \mathbb{R}^{H \times W \times 3},
\end{equation}
where $I_t^{\uparrow}$ is the result of bilinear interpolation of $I_t$ and $\smallfrown[4pt]$ denotes the concatenation operation along the channel dimension. $\mathcal{F}$ is the backbone of upsampling network parameterized by $\delta $, which is mainly composed of encoder, decoder and several residual blocks. More details about the architecture could be found in the supplementary material.

We optimize this guided upsampling network by minimizing  $\ell_{1}$ loss between $I_{pred}$ and corresponding ground-truth $I$: 
\begin{equation}
    L_{\ell_{1}}=\mathbb{E}(\left\|I_{pred}-I\right\|_{1}).
\end{equation}
To generate more realistic details, extra adversarial loss is also involved in the training process, specifically,
\begin{equation}
    L_{adv} =\mathbb{E}\left[\log \left(1-\mathcal{D}_\omega \left(I_{pred}\right)\right)\right]+\mathbb{E}\left[\log \mathcal{D}_\omega \left(I\right)\right],
\end{equation}
where $\mathcal{D}$ is the discriminator parameterized by $\omega$. We jointly train upsampling network $\mathcal{F}$ and discriminator $\mathcal{D}$ through solving the following optimization,
\begin{equation}
    \min _{\mathcal{F}} \max _{\mathcal{D}}L_{upsample}(\delta,\omega )=\alpha_{1}L_{\ell_{1}}+\alpha_{2}L_{adv}.
\end{equation}
The loss weights are set to $\alpha_{1}=1.0$ and $\alpha_{2}=0.1$ in all experiments. We also observe that involving instance normalization (IN)~\cite{ulyanov2016instance} will cause color inconsistency and severe artifacts during optimization. Therefore we remove all IN in the upsampling network.

\begin{figure*}[t!]
    \begin{center}
    \includegraphics[width=1.0\linewidth]{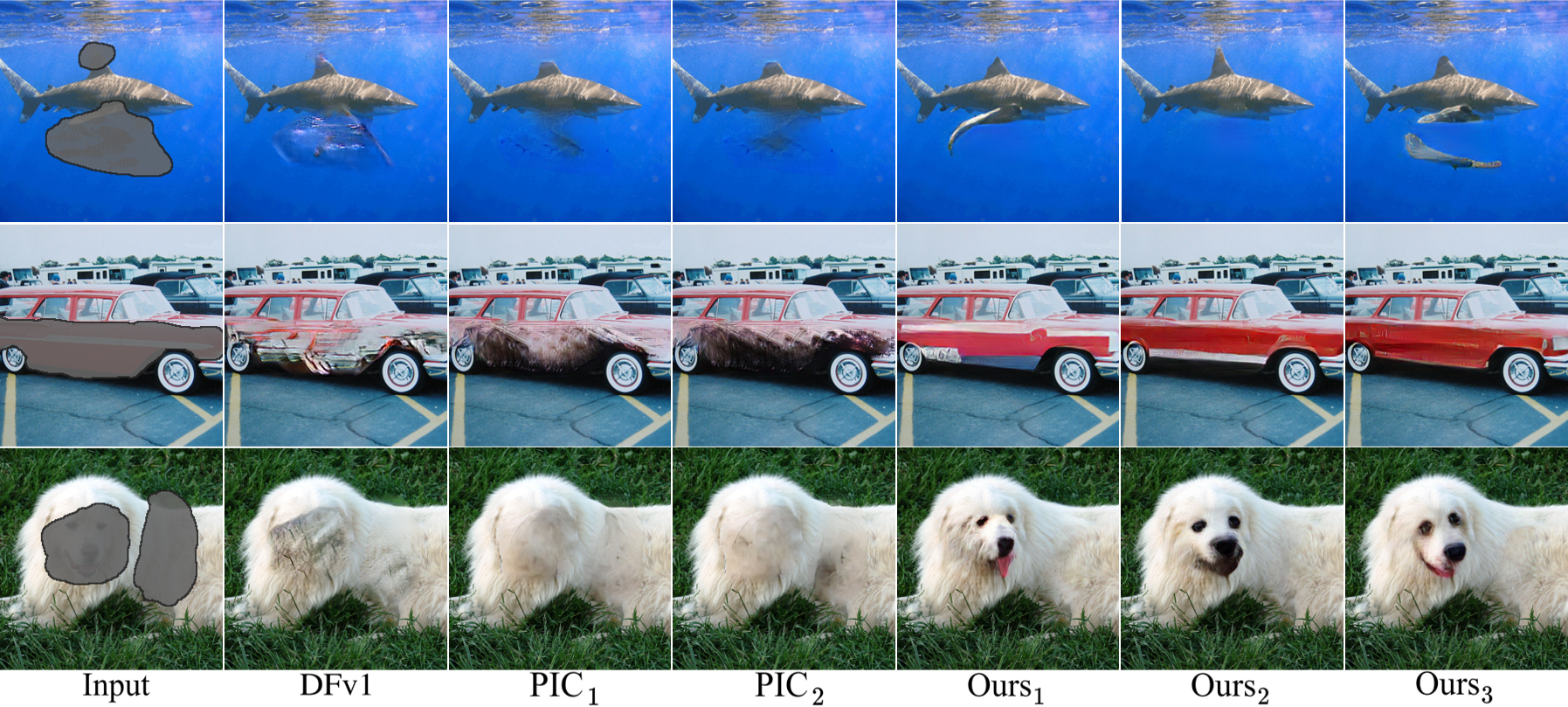}
    \end{center}
    \vspace{-1.4em}
    \caption{\textbf{Qualitative comparison with state-of-the-art methods on ImageNet dataset.} More qualitative examples are shown in supplementary materials.}
    \label{figure:qualititative_ImageNet}
    \vspace{-0.8em}
\end{figure*}

\section{Experiments} 
We present the details of implementation in Sec.~\ref{sec.4.1}, subsequently evaluate (Sec.~\ref{sec.4.2}) and delve into (Sec.~\ref{sec.4.3}) the proposed transformer-based image completion method. The pluralistic image completion experiments are conducted at $256 \times 256$ resolution on three datasets including \textit{FFHQ}~\cite{ffhq}, \textit{Places2}~\cite{zhou2017places} and \textit{ImageNet}~\cite{russakovsky2015imagenet}. We preserve 1K images from the whole \textit{FFHQ} for testing, and use the original common training and test splits in rest datasets. The diversified irregular mask dataset provided by PConv~\cite{liu2018image} is employed for both training and evaluation.

\subsection{Implementation Details} \label{sec.4.1}
We control the scale of transformer architecture by balancing the representation ability and size of dataset concurrently. The discrete sequence length $\mathbb{L}$ is set to $48\times48$ for \textit{FFHQ}. Limited by the computational resources, we decrease feasible $\mathbb{L}$ to $32\times32$ on large-scale datasets \textit{Places2} and \textit{ImageNet}.  The detailed configurations of different transformer models are attached in supplementary material.

We use $8$ Tesla V100 GPUs for \textit{FFHQ} with batch size 64, $4\times8$ Tesla V100 GPUs for \textit{Places2} and \textit{ImageNet} with batch size 256 to train the transformers until convergence. We optimize the network parameters using AdamW~\cite{adamw} with $\beta_1=0.9$ and $\beta_2=0.95$. The learning rate is warmed up from 0 to $3\text{e-}4$ linearly in the first epoch, then decays to 0 via cosine scheduler in rest iterations. No extra weight decay and dropout strategy are employed in the model. To train the guided upsampling network we use Adam~\cite{kingma2014adam} optimizer with fixed learning rate $1\text{e-}4$, $\beta_1=0.0$ and $\beta_2=0.9$. During optimization, the weights of different loss terms are set to fixed value described in Sec.~\ref{sec3.2} empirically.

\subsection{Results} \label{sec.4.2}
Our method is compared against the following state-of-the-art inpainting algorithms: $\text{Edge-Connect}_{\text{ICCV'19}}\text{(EC)}$ \cite{nazeri2019edgeconnect}, $\text{DeepFillv2}_{\text{ICCV'19}}$(DFv2)~\cite{yu2019free}, $\text{MED}_{\text{ECCV'20}}$ \cite{liu2020rethinking} and $\text{PIC}_{\text{CVPR'19}}$ \cite{zheng2019pluralistic} using the official pre-trained models. We also fully train these models on \textit{FFHQ} dataset for fair comparison.

\noindent\textbf{Qualitative Comparisons} ~~ We qualitatively compare the results with other baselines in this section. We adopt the sampling strategy introduced in Sec.~\ref{sec3.1} with $\mathcal{K}$=50 to generate 20 solutions in parallel, then select the top 6 results ranked by the discriminator score of the upsampling network following PIC~\cite{zheng2019pluralistic}. All reported results of our method are direct outputs from the trained models without extra post-processing steps. 

We show the results on the \textit{FFHQ} and \textit{Places2} datasets in Figure.~\ref{figure:qualititative_face_place}.  Specifically, EC~\cite{nazeri2019edgeconnect} and MED~\cite{liu2020rethinking} could generally produce the basic components of missing regions, but the absence of texture details makes their results nonphotorealistic. DeepFillv2~\cite{yu2019free}, which is based on a multi-stage restoration framework, could generate sharp details. However, severe artifacts appear when mask regions are relatively large. Besides, their method could only produce a single solution for each input. As the state-of-the-art diversified image inpainting method, PIC~\cite{zheng2019pluralistic} tends to generate over-smooth contents and strange patterns, meanwhile the semantic-reasonable variation is limited to a small range. Compared to these methods, ours is superior in both photo-realism and diversity. We further show the comparison with $\text{DeepFillv1}_{\text{CVPR'18}}\text{(DFv1)}$ \cite{Yu-cvpr18-attentioninpainting8} and PIC~\cite{zheng2019pluralistic} on ImageNet dataset in Figure.~\ref{figure:qualititative_ImageNet}. In this challenging setting, full CNN-based methods could not understand the global context well, resulting in unreasonable completion. When meeting large masks, they even could not maintain the accurate structure as shown in the second row of Figure.~\ref{figure:qualititative_ImageNet}. In comparison, our method gives superior results, which demonstrates the exceptional generalization ability of our method on large-scale datasets.

\begin{table*}[!t]
\begin{center}
\begin{tabular}{CCCCCCCCCC}
\hline
\multicolumn{2}{c}{\textsc{Dataset}}                                                       & \multicolumn{4}{c}{\textsc{Ffhq}~\cite{ffhq}}                                                                                                                                    & \multicolumn{4}{c}{\textsc{Places2}~\cite{zhou2017places}}                                                                                                                                 \\ \hline
\multicolumn{1}{l|}{Method}    & \multicolumn{1}{c|}{Mask Ratio}                  &  PSNR$\uparrow$                          & SSIM$\uparrow$                           & MAE$\downarrow$                           & FID$\downarrow$                           & PSNR$\uparrow$                            & SSIM$\uparrow$                           & MAE$\downarrow$                           & FID$\downarrow$                           \\ \hline
\multicolumn{1}{l|}{DFv2~\cite{yu2019free}}      & \multicolumn{1}{c|}{}                            & 25.868                                  & 0.922                                  & 0.0231                                  & 16.278                                  & 26.533                                  & 0.881                                  & 0.0215                                   & 24.763                                  \\
\multicolumn{1}{l|}{EC~\cite{nazeri2019edgeconnect}}        & \multicolumn{1}{c|}{}                            &   26.901                                & 0.938                                  & 0.0209                                  &             14.276                   &   26.520                                      &     0.880                                   &                      0.0220                   &         25.642                                      \\
\multicolumn{1}{l|}{PIC~\cite{zheng2019pluralistic}}       & \multicolumn{1}{c|}{}                            & 26.781                                  & 0.933                                  & 0.0215                                  & 14.513                                  & 26.099                                  & 0.865                                  & 0.0236                                  & 26.393                                  \\
\multicolumn{1}{l|}{MED~\cite{liu2020rethinking}}       & \multicolumn{1}{c|}{}                            & 26.325                                  & 0.922                                  & 0.0230                                  & 14.791                                  & 26.469                                  & 0.877                                  & 0.0224                                  & 26.977                                  \\
\multicolumn{1}{l|}{Ours} & \multicolumn{1}{c|}{}                            & \cellcolor[HTML]{EFEFEF}27.922          & \cellcolor[HTML]{EFEFEF}0.948          & \cellcolor[HTML]{EFEFEF}0.0208          & \cellcolor[HTML]{EFEFEF}10.995          & \cellcolor[HTML]{EFEFEF}26.503          & \cellcolor[HTML]{EFEFEF}0.880          & \cellcolor[HTML]{EFEFEF}0.0244          & \cellcolor[HTML]{EFEFEF}21.598          \\
\multicolumn{1}{l|}{$\text{Ours}^{\dagger}$}  & \multicolumn{1}{c|}{\multirow{-6}{*}{20\%-40\%}} & \cellcolor[HTML]{EFEFEF}\textbf{28.242} & \cellcolor[HTML]{EFEFEF}\textbf{0.952} & \cellcolor[HTML]{EFEFEF}\textbf{0.0155} & \cellcolor[HTML]{EFEFEF}\textbf{10.515} & \cellcolor[HTML]{EFEFEF}\textbf{26.712} & \cellcolor[HTML]{EFEFEF}\textbf{0.884} & \cellcolor[HTML]{EFEFEF}\textbf{0.0198} & \cellcolor[HTML]{EFEFEF}\textbf{20.431} \\ \hline
\multicolumn{1}{l|}{DFv2~\cite{yu2019free}}      & \multicolumn{1}{c|}{}                            & 21.108                                  & 0.802                                  & 0.0510                                   & 28.711                                  & 22.192                                  & 0.729                                  & 0.0440                                   & 39.017                                  \\
\multicolumn{1}{l|}{EC~\cite{nazeri2019edgeconnect}}        & \multicolumn{1}{c|}{}                            & 21.368                                  & 0.780                                  & 0.0510                                  &          30.499                         &   22.225                                      &             0.731                           &               0.0438                          &    39.271                                     \\
\multicolumn{1}{l|}{PIC~\cite{zheng2019pluralistic}}       & \multicolumn{1}{c|}{}                            & 21.723                                  & 0.811                                  & 0.0488                                  & 25.031                                  & 21.498                                  & 0.680                                  & 0.0507                                  & 49.093                                  \\
\multicolumn{1}{l|}{MED~\cite{liu2020rethinking}}       & \multicolumn{1}{c|}{}                            & 20.765                                  & 0.763                                  & 0.0592                                  & 34.148                                  & 22.271                                  & 0.717                                  & 0.0457                                  & 45.455                                  \\
\multicolumn{1}{l|}{Ours} & \multicolumn{1}{c|}{}                            & \cellcolor[HTML]{EFEFEF}22.613          & \cellcolor[HTML]{EFEFEF}0.845          & \cellcolor[HTML]{EFEFEF}0.0445          & \cellcolor[HTML]{EFEFEF}\textbf{20.024} & \cellcolor[HTML]{EFEFEF}22.215          & \cellcolor[HTML]{EFEFEF}0.724          & \cellcolor[HTML]{EFEFEF}0.0431          & \cellcolor[HTML]{EFEFEF}\textbf{33.853} \\
\multicolumn{1}{l|}{$\text{Ours}^{\dagger}$}  & \multicolumn{1}{c|}{\multirow{-6}{*}{40\%-60\%}} & \cellcolor[HTML]{EFEFEF}\textbf{23.076} & \cellcolor[HTML]{EFEFEF}\textbf{0.864} & \cellcolor[HTML]{EFEFEF}\textbf{0.0371} & \cellcolor[HTML]{EFEFEF}20.843          & \cellcolor[HTML]{EFEFEF}\textbf{22.635} & \cellcolor[HTML]{EFEFEF}\textbf{0.739} & \cellcolor[HTML]{EFEFEF}\textbf{0.0401} & \cellcolor[HTML]{EFEFEF}34.206          \\ \hline
\multicolumn{1}{l|}{DFv2~\cite{yu2019free}}      & \multicolumn{1}{c|}{}                            & 24.962                                  & 0.882                                  & 0.0310                                   & 19.506                                  & 25.692                                  & 0.834                                  & 0.0280                                   & 29.981                                  \\
\multicolumn{1}{l|}{EC~\cite{nazeri2019edgeconnect}}        & \multicolumn{1}{c|}{}                            & 25.908                                  & 0.882                                  & 0.0301                                  &           17.039                        &   25.510                                      &           0.831                             &                  0.0293                       &    30.130                                      \\
\multicolumn{1}{l|}{PIC~\cite{zheng2019pluralistic}}       & \multicolumn{1}{c|}{}                            & 25.580                                  & 0.889                                  & 0.0303                                  & 17.364                                  & 25.035                                  & 0.806                                  & 0.0315                                  & 33.472                                  \\
\multicolumn{1}{l|}{MED~\cite{liu2020rethinking}}       & \multicolumn{1}{c|}{}                            & 25.118                                  & 0.867                                  & 0.0349                                  & 19.644                                  & 25.632                                  & 0.827                                  & 0.0291                                  & 31.395                                  \\
\multicolumn{1}{l|}{Ours} & \multicolumn{1}{c|}{}                            & \cellcolor[HTML]{EFEFEF}26.681          & \cellcolor[HTML]{EFEFEF}0.910          & \cellcolor[HTML]{EFEFEF}0.0292          & \cellcolor[HTML]{EFEFEF}14.529          & \cellcolor[HTML]{EFEFEF}25.788          & \cellcolor[HTML]{EFEFEF}0.832          & \cellcolor[HTML]{EFEFEF}0.0267          & \cellcolor[HTML]{EFEFEF}\textbf{25.420} \\
\multicolumn{1}{l|}{$\text{Ours}^{\dagger}$}  & \multicolumn{1}{c|}{\multirow{-6}{*}{Random}}    & \cellcolor[HTML]{EFEFEF}\textbf{27.157} & \cellcolor[HTML]{EFEFEF}\textbf{0.922} & \cellcolor[HTML]{EFEFEF}\textbf{0.0223} & \cellcolor[HTML]{EFEFEF}\textbf{14.039} & \cellcolor[HTML]{EFEFEF}\textbf{25.982} & \cellcolor[HTML]{EFEFEF}\textbf{0.839} & \cellcolor[HTML]{EFEFEF}\textbf{0.0254} & \cellcolor[HTML]{EFEFEF}25.985          \\ \hline
\end{tabular}
\caption{\textbf{Quantitative results on FFHQ and Places2 datasets with different mask ratios.} $\text{Ours}$: Default top-50 sampling. $\text{Ours}^{\dagger}$: Top-1 sampling.}
\label{tab:quantitative_1}
\end{center}
\vspace{-1em}
\end{table*}

\begin{table}[!t]
\vspace{-1.5em}
\scalebox{0.9}{
\begin{tabular}{l|c|cccc}
\hline
Method & Mask Ratio                  & PSNR$\uparrow $                           & SSIM$\uparrow   $                        & MAE$\downarrow  $                         & FID$\downarrow  $                         \\ \hline
PIC~\cite{zheng2019pluralistic}    &                             & 24.010                                  & 0.867                                  & 0.0319                                  & 47.750                                  \\
Ours   & \multirow{-2}{*}{20\%-40\%} & \cellcolor[HTML]{EFEFEF}\textbf{24.757} & \cellcolor[HTML]{EFEFEF}\textbf{0.888} & \cellcolor[HTML]{EFEFEF}\textbf{0.0263} & \cellcolor[HTML]{EFEFEF}\textbf{28.818} \\ \hline
PIC~\cite{zheng2019pluralistic}    &                             & 18.843                                  & 0.642                                  & 0.0756                                  & 101.278                                 \\
Ours   & \multirow{-2}{*}{40\%-60\%} & \cellcolor[HTML]{EFEFEF}\textbf{20.135} & \cellcolor[HTML]{EFEFEF}\textbf{0.721} & \cellcolor[HTML]{EFEFEF}\textbf{0.0585} & \cellcolor[HTML]{EFEFEF}\textbf{59.486} \\ \hline
PIC~\cite{zheng2019pluralistic}    &                             & 22.711                                  & 0.791                                  & 0.0462                                  & 59.428                                  \\
Ours   & \multirow{-2}{*}{Random}    & \cellcolor[HTML]{EFEFEF}\textbf{23.775} & \cellcolor[HTML]{EFEFEF}\textbf{0.835} & \cellcolor[HTML]{EFEFEF}\textbf{0.0358} & \cellcolor[HTML]{EFEFEF}\textbf{35.842} \\ \hline
\end{tabular}}
\caption{\textbf{Quantitative comparison with PIC on ImageNet dataset.}}
\label{tab: quantitative_2}
\vspace{-1.5em}
\end{table}

\noindent\textbf{Quantitative Comparisons} ~~ We numerically compare our method with other baselines in Table.~\ref{tab:quantitative_1} and Table.~\ref{tab: quantitative_2}. The peak signal-to-noise ratio (PSNR), structural similarity index (SSIM), and relative $\ell_{1}$ (MAE) are used to compare the low-level differences between the recovered output and the ground truth, which is more suitable to measure the mask setting of small ratio. To evaluate the larger area missing, we adopt Fr\'echet Inception Distance (FID)~\cite{FID}, which calculates the feature distribution distance between completion results and natural images. Since our method could produce multiple solutions, we need to find one examplar to calculate the mentioned metrics. Unlike PIC~\cite{zheng2019pluralistic}, which selects the results with high ranking discriminator scores for each sample, here we directly provide stochastic sampling results while $\mathcal{K}$=50 to demonstrate its generalization capability.  Besides, we also provide deterministic sampling results given $\mathcal{K}$=1 in Table.~\ref{tab:quantitative_1}. It can be seen that our method with top-1 sampling achieves superior results compared with other competitors in almost all metrics. And in the case of relatively large mask regions, top-50 sampling leads to slightly better FID scores. On the ImageNet dataset, as shown in Table.~\ref{tab: quantitative_2}, our method outperforms PIC by a considerable margin, especially in FID metrics (more than 41.2 for large masks).

\begin{figure}[t!]
    \begin{center}
    \includegraphics[width=1.0\linewidth]{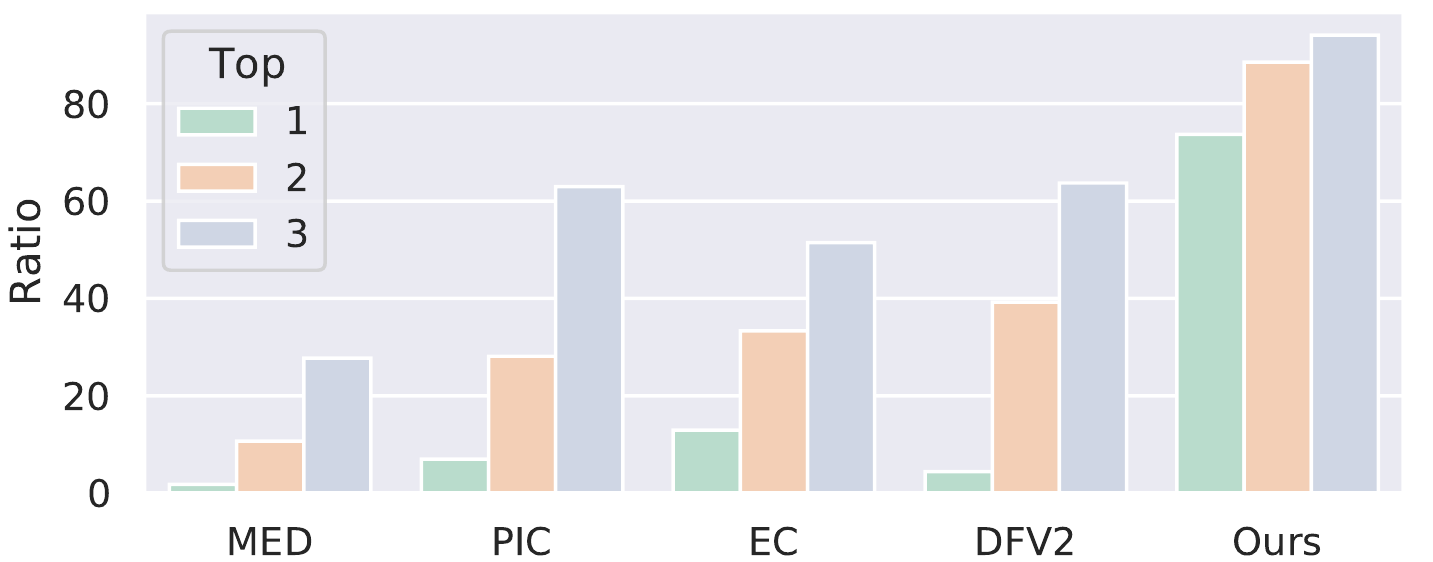}
    \end{center}
    \vspace{-1.4em}
    \caption{\textbf{Results of user study.}}
    \label{figure:user_study}
    \vspace{-2.0em}
\end{figure}

\noindent\textbf{User Study} ~~ To better evaluate the subjective quality, we further conduct a user study to compare our method with other baselines. Specifically, we randomly select 30 masked images from the test set. For a test image, we use each method to generate one completion result and ask the participant to rank five results from the highest photorealism to the lowest photorealism. We have collected answers from 28 participants and calculate the ratios of each method being selected as top 1,2,3, with the statistics shown in Figure.~\ref{figure:user_study}. Our method is 73.70\% more likely to be picked as the first rank, demonstrating its clear advantage.

\begin{figure}%
    \vspace{-2.5em}
    \centering
    \subfloat{{\includegraphics[width=4.2cm]{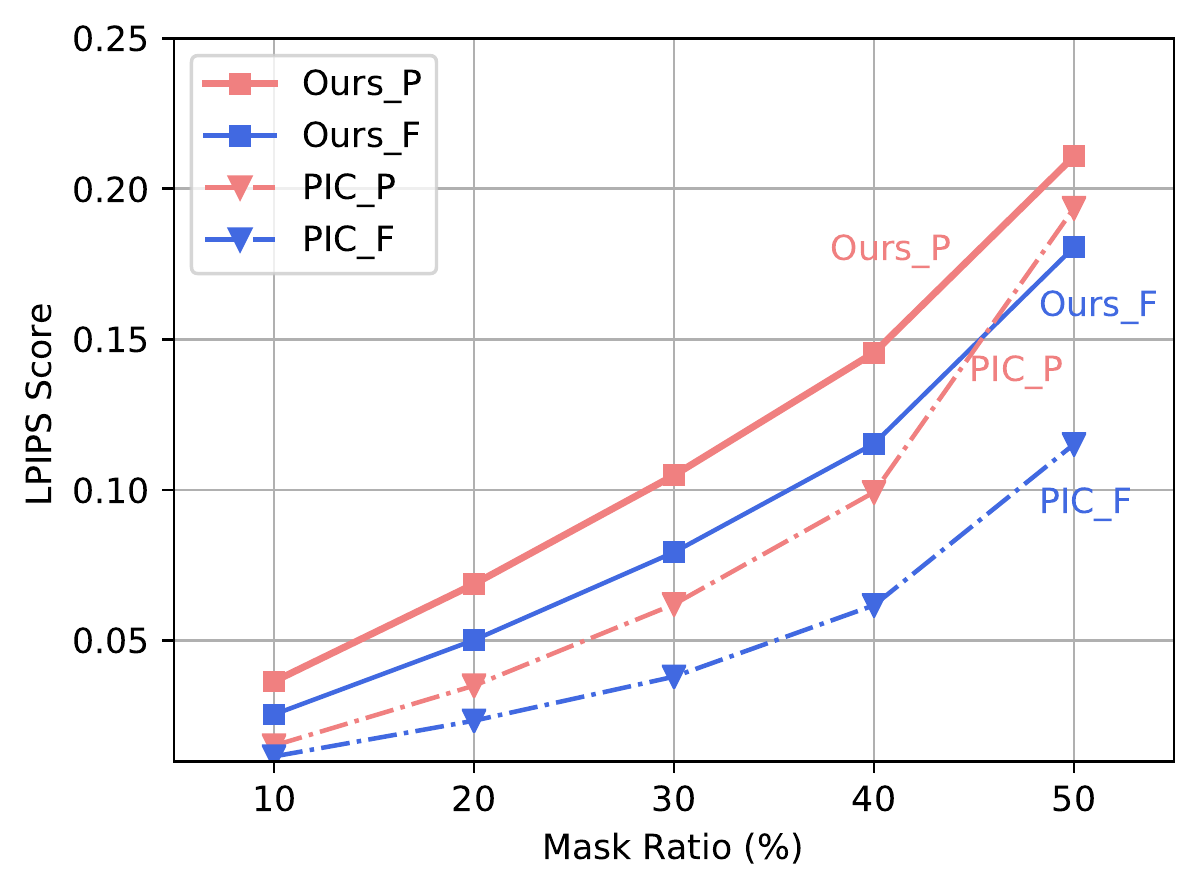} }}%
    \subfloat{{\includegraphics[width=4.2cm]{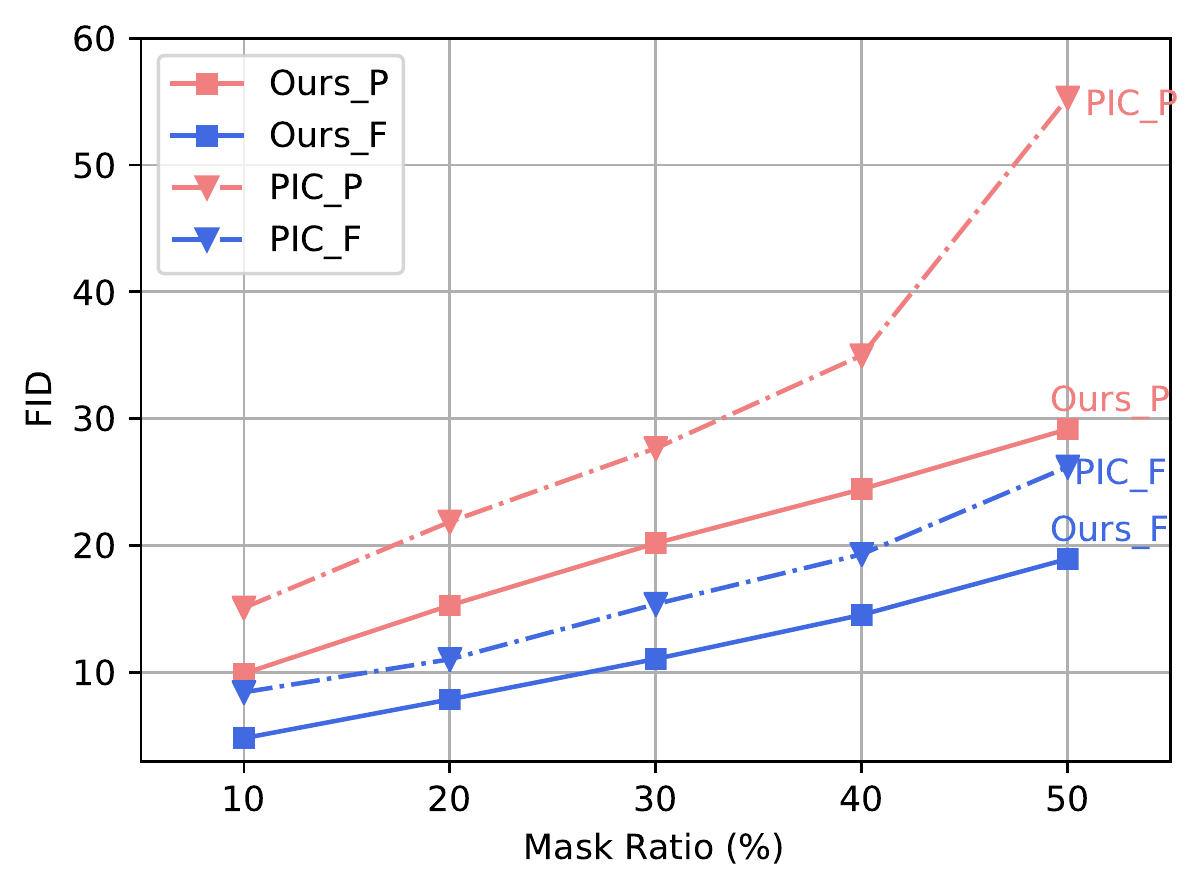} }}%
    \caption{\textbf{Left:} Diversity curve. \textbf{Right:} FID curve. \textbf{P} and \textbf{F} denote \textit{Places2} and \textit{FFHQ} dataset respectively.}%
    \label{fig:LPIPS}%
        \vspace{-2.0em}
\end{figure}

\begin{figure*}[!t]
    \begin{center}
    \includegraphics[width=1.0\linewidth]{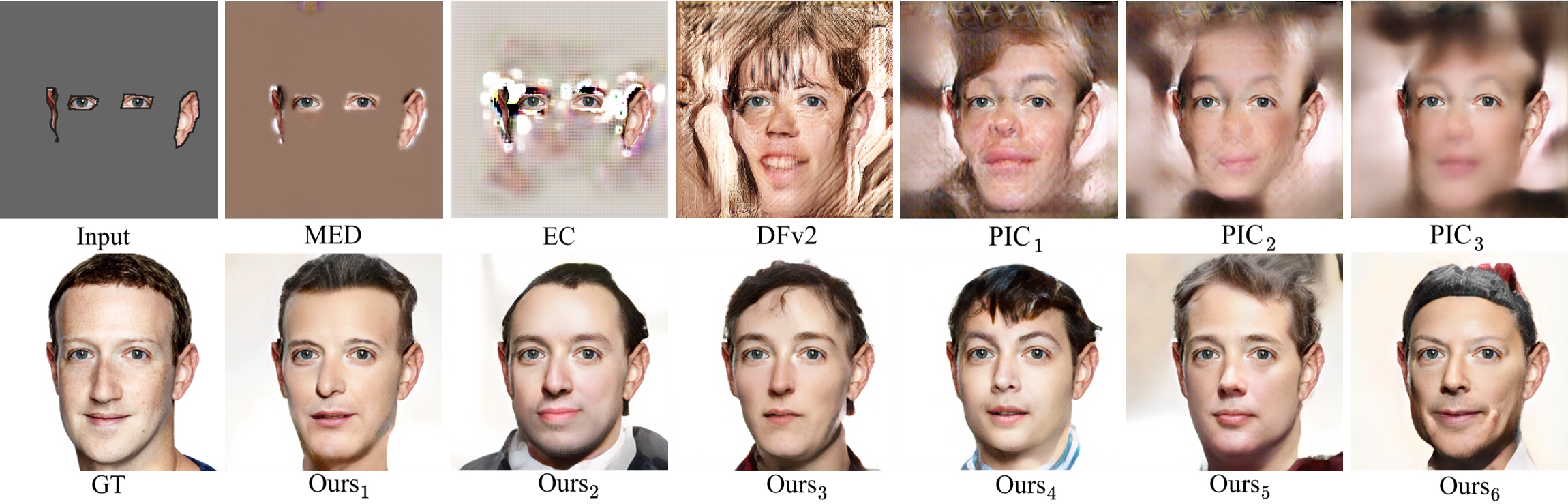}
    \end{center}
    \vspace{-1.4em}
    \caption{\textbf{Image completion results of large-scale masks.} It could be noted that all the compared baselines struggle to generate plausible contents.}
    \label{figure:large-scale}
    \vspace{-1.0em}
\end{figure*}

\subsection{Analysis} \label{sec.4.3}
\noindent\textbf{Diversity.} ~~ We calculate the average LPIPS distance~\cite{zhang2018unreasonable} between pairs of randomly-sampled outputs from the same input to measure the completion diversity following Zhu \emph{et al.}~\cite{zhu2017toward}. Specifically, we leverage 1K input images and sample 5 output pairs per input in different mask ratios. And the LPIPS is computed based on the deep features of VGG~\cite{simonyan2014very} model pre-trained on ImageNet. The diversity scores are shown in Figure.~\ref{fig:LPIPS}. Since the completion with diverse but meaningless contents will also lead to high LPIPS scores, we simultaneously provide the FID score of each level computed between the whole sampled results (10K) and natural images in the right part of Figure.~\ref{fig:LPIPS}. Our method achieves better diversity in all cases. Besides, in the max mask ratio of Places2, although PIC~\cite{zheng2019pluralistic} approximates the diversity of our method, the perceptual quality of our completion outperforms PIC~\cite{zheng2019pluralistic} by a large margin.

\begin{figure}[t!]
\vspace{-0.3em}
    \begin{center}
    \includegraphics[width=1.0\linewidth]{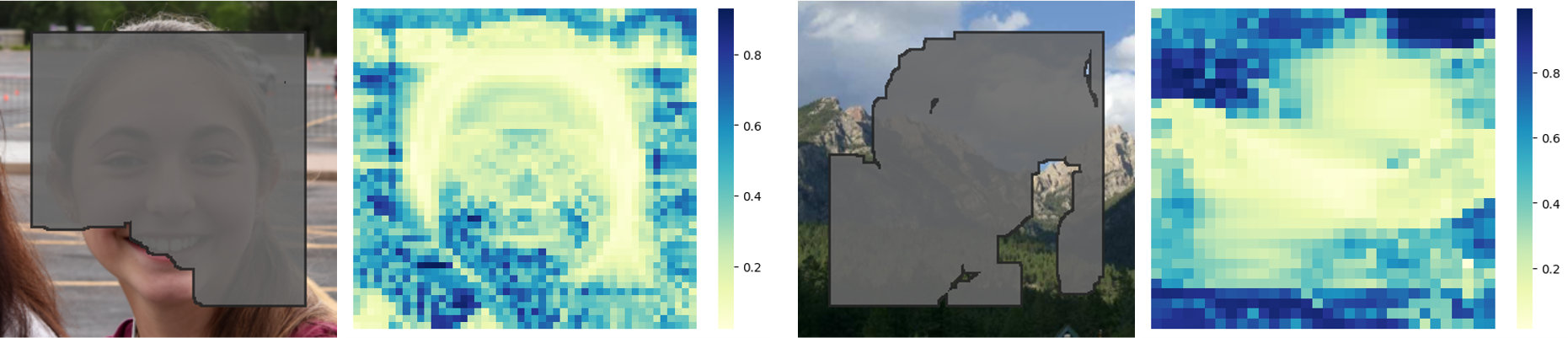}
    \end{center}
    \vspace{-1.4em}
    \caption{\textbf{Visualization of probability map generated by transformer.} Higher confidence denotes lower uncertainty.}
    \label{figure:uncertainty}
    \vspace{-1.5em}
\end{figure}

\noindent\textbf{Robustness for the completion of extremely large holes.} ~~ To further understand the ability of the transformer, we conduct extra experiments on the setting of extremely large holes, which means only very limited pixels are visible. Although both the transformer and upsampling network are only trained using a dataset of PConv~\cite{liu2018image} (max mask ratio 60\%), our method generalizes fairly well to this difficult setting. In Figure.~\ref{figure:large-scale}, almost all the baselines fail with large missing regions, while our method could produce high-quality and diversified completion results.

\begin{figure}[t!]
 \vspace{-0.5em}
    \begin{center}
    \includegraphics[width=1.0\linewidth]{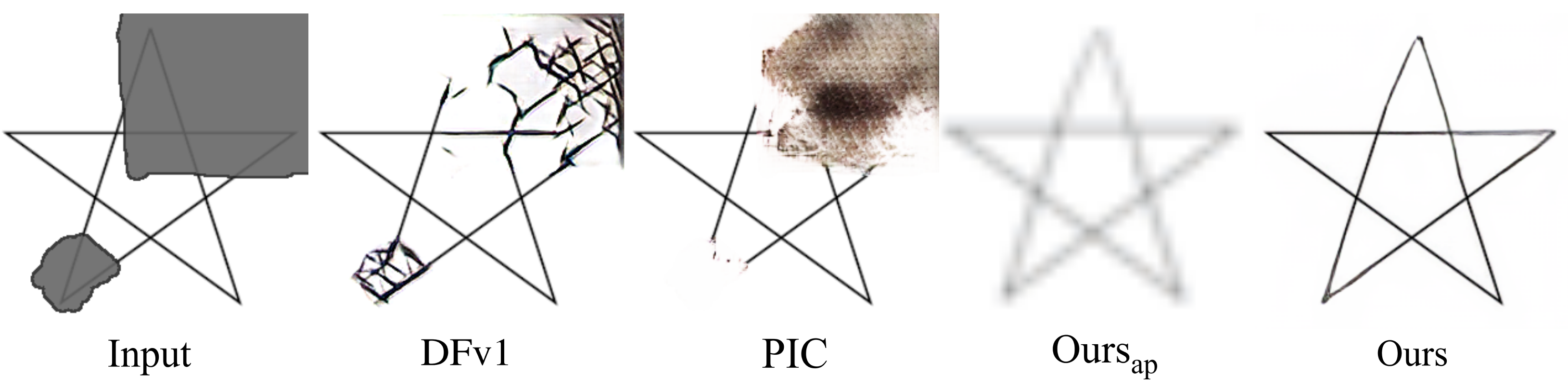}
    \end{center}
    \vspace{-1.8em}
    \caption{\textbf{Completion of basic geometric shape.} All compared models are trained on ImageNet. $\text{Ours}_{\text{ap}}$: Appearance prior reconstructed from transformer. }
    \label{figure:GEOMETRY}
    \vspace{-1.4em}
\end{figure}

\noindent\textbf{If the transformer could better understand global structure than CNN?} ~~ To answer this question, we conduct the completion experiments on some geometric primitives. Specifically, we ask our transformer-based method and other full CNN methods, \ie DeepFillv1~\cite{Yu-cvpr18-attentioninpainting8} and PIC~\cite{zheng2019pluralistic}, trained on ImageNet to recover the missing parts of pentagram shape in Figure.~\ref{figure:GEOMETRY}. As expected, all full CNN methods fail to reconstruct the missing shape, which may be caused by the locality of the convolution kernel. In contrast, the transformer could easily reconstruct the right geometry in low-dimensional discrete space. Based on such accurate appearance prior, the upsampling network could more effectively render the original resolution results finally.

\noindent\textbf{Visualization of probability map.} ~~ Intuitively, since the contour of missing regions is contiguous to existing pixels, the completion confidence should gradually decrease from the mask boundary to the interior region. The lower confidence corresponds to more diversified results. To verify this point, we plot the probability map in Figure.~\ref{figure:uncertainty} , where each pixel denotes the maximum probability of visual vocabulary generated by the transformer. And we have some interesting observations: 1) In the right part of Figure.~\ref{figure:uncertainty}, the uncertainty is indeed increasing from outside to inside. 2) For the portrait completion example, the uncertainty of face regions is overall lower than hair parts. The underlying reason is the seen parts of the face constrain the diversity of other regions to some degree. 3) The probability of the right cheek of the portrait example is highest among the rest mask regions, which indicates that the transformer captures the symmetric property.

\section{Concluding Remarks} 
There is a long-existing dilemma in the image completion area to achieve both enough diversity and photorealistic quality. Existing attempts mostly optimize the variational lower-bound through a full CNN architecture, which not only limits the generation quality but also is difficult to render natural variations. In this paper, we first propose to bring the best of both worlds: structural understanding capability and pluralism support of transformers, and local texture enhancement and efficiency of CNNs, to achieve high-fidelity free-form pluralistic image completion. Extensive experiments are conducted to demonstrate that the superiority of our method compared with state-of-the-art fully convolutional approaches, including large performance gain on regular evaluations setting, more diversified and vivid results, and exceptional generalization ability on large-scale masks and datasets.

{\small
\bibliographystyle{ieee_fullname}
\bibliography{egbib}
}

\end{document}